\documentclass[letterpaper, 10 pt, conference]{ieeeconf}

\IEEEoverridecommandlockouts                              

\usepackage{times,enumerate} 
\usepackage[usenames,dvipsnames,svgnames,table]{xcolor}

\usepackage{subfloat}
\usepackage{hyperref}

\usepackage{graphicx}
\usepackage{setspace}
\usepackage{bbm}
\usepackage{mathdots,mathrsfs}
\usepackage{amssymb,latexsym,amsfonts,amsmath,cite,comment,relsize}
\usepackage{stmaryrd}
\usepackage{caption}
\usepackage[dvips]{psfrag}
\usepackage{setspace}
\usepackage{amsmath}
\usepackage{url}
\usepackage{soul}
\usepackage{blindtext}
\usepackage{graphicx}
\usepackage{lipsum}

\usepackage{adjustbox}
\usepackage{tikz}
\usepackage{pgfplots}
\pgfplotsset{every tick label/.append style={font=\small}}

\usepackage{resizegather}

\usepackage{relsize}
\usetikzlibrary{patterns}

\usepackage{epsfig} 
\usepackage{subfig}
\usepackage{algorithm,algpseudocode}

\newtheorem{example}{Example}

\newtheorem{remark}{Remark}

\tikzstyle{vertex}=[circle, draw, inner sep=0pt, minimum size=6pt]

\usepackage[sans]{dsfont}
\usepackage[font={small,it}]{caption}

\newcommand{\suchthat}{\;\ifnum\currentgrouptype=16 \middle\fi|\;}

\newcommand{\R}{\mathbb{R}}
\newcommand{\Z}{\mathbb{Z}}

\newcommand{\G}{\mathcal{G}}
\newcommand{\E}{\mathcal{E}}
\newcommand{\V}{\mathcal{V}}

\usepackage{algorithm,algpseudocode}

\title{\LARGE \bf
	Multi-Agent Image Classification via Reinforcement Learning  
}

\author{Hossein K. Mousavi, Mohammadreza Nazari, Martin Tak\'a\u{c}, and Nader Motee$^{*}$
	\thanks{*The first and second author have equal contribution to this work.}
	\thanks{H.K.M. and N.M. are with the department of Mechanical Engineering and Mechanics, Lehigh University, Bethlehem, PA 18015, USA \newline {\tt\small \{mousavi, motee\}@lehigh.edu}. }
	\thanks{M.N. and M.T. are with the department of Industrial and Systems Engineering, Lehigh University, Bethlehem, PA 18015, USA {\tt\small \{mrza.nazari, takac.mt\}@gmail.com}. }
	\thanks{This work was supported in parts by the NSF CAREER ECCS-1454022, NSF CCF-1618717, NSF CMMI-1663256, NSF CCF-1740796, ONR YIP N00014-16-1-2645,
		and ONR N00014-19-1-2478. 	
	}
}

\usepackage{xargs}                      

\begin{document}
	
	\maketitle
	\thispagestyle{empty}
	\pagestyle{empty}

	\begin{abstract}  
		We investigate a classification problem using multiple mobile agents 
		 capable of collecting (partial) pose-dependent observations of an unknown environment. The objective is to classify an image  over a finite time horizon. We propose a network architecture on how agents should form a local belief, take local actions, and extract relevant features   from their raw partial observations. Agents are allowed to exchange information with their neighboring agents   to update their own beliefs. It is shown how reinforcement learning techniques can be utilized to achieve decentralized implementation of the classification problem  by running a decentralized consensus protocol.  Our experimental results on the MNIST handwritten digit dataset demonstrates the effectiveness of our proposed framework.
	\end{abstract}


	\section{Introduction}
	
	With the  {rising interest in }   the Internet of Things (IoT), the demand for {design of autonomous agents that are capable of cooperation is increasing}. The interconnected robots will be major players in the future, accomplishing many duties in industrial automation\cite{bahrin2016industry}, military support\cite{yushi2012study}, and health-care\cite{bhatt2017internet}. {In many of these applications, a major issue is that every agent has limited sensing capabilities, and therefore, may not have sufficient information for accomplishing a complex task. One way to mitigate this shortcoming is to let the task to be solved collectively by multiple agents. {In the context of machine learning,} this means that the agents need not only to learn through individual interaction with their environment but also they can {learn} from each others' experiences using {communication.}
	
	  In several {machine learning} applications, the problem suffers from the high-dimension of the feature space, which may {render} the learning process inefficient. One may list facial element recognition, genome disorder identification, or fault detection as instances of problems facing these issues. In these examples, the main challenge is that a large portion of the input data might be irrelevant to the task. One possibility is to pre-process the data to filter the irrelevant pieces of information.  However, this might be challenging or cause data inaccuracy.  In this paper, we propose an approach that may provide an alternative mechanism for complexity reduction:  we translate the classification task into a multi-agent reinforcement learning setting, where the space of observations per agent has a comparatively lower dimension. 
	  
	  	\begin{figure}[t]
	  	\centering
	  	\includegraphics*[width=8.2cm]{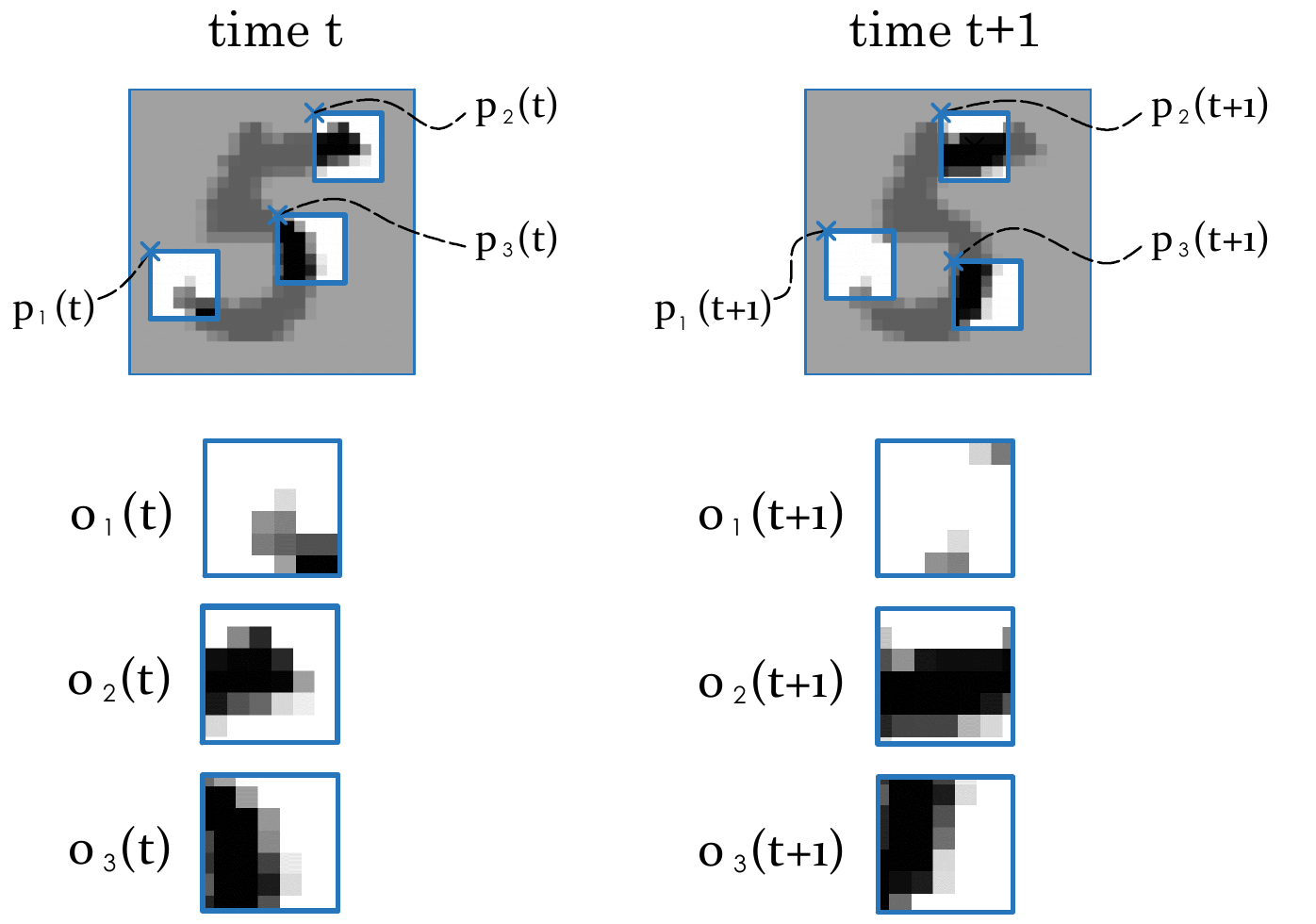}
	  	\caption{{This example illustrates why classification through local observations of an agent is challenging. } The observations { made} by $3$ agents at two consecutive time instants have been magnified{, where the spatial variables dictate the observations}.  This highlights the need for {a communication and memory mechanisms.}}
	  	\label{fig:five}
	  \end{figure}

	{
		
		In this work, we study the multi-agent image classification problem within an unknown environment. The setup consists of an environment, in which multiple homogeneous agent,s each with a partial observation of the environment, are collaborating  to do a classification task. To explore the environment efficiently, the agents need to learn how to optimally traverse the environment and receive new observations   through re-positioning. {Moreover,} they are    capable of establishing autonomous communication  to update their beliefs. This is motivated by the idea that a more experienced teammate could provide or explain hints, which can be used to { facilitate the perception. We are interested in maximizing a long-term {collective} reward that pushes the agents  to effectively coordinate and cooperate in order to correctly classify the image. Our goal in this work is to {approach this classification task through an end-to-end} co-design {of} decentralized data-processing, communication, action planning, and  prediction modules on each agent (see Fig. \ref{fig:five} for an example).

		For solving this problem, we formulate it within a multi-agent reinforcement learning   framework and propose a policy gradient, which can effectively optimize the agents' behavior. In contrast to the vanilla policy gradient  settings, our framework introduces a differentiable reward rather instead of a reward that is independent of the network parameters. The proposed mechanism enables the policy gradient algorithm to not only maximize the probability of generating desirable outcomes, but also to explicitly increase the rewards. The mathematical derivation of the latter argument is given in Section \ref{sec:three}, which generalizes the policy gradient approach for the case of differentiable rewards.
		
		
		We have demonstrated a {descent} performance of the proposed framework in the MNIST classification task.  We can correctly classify 88\% of the testing dataset by using two agents, each only with 2$\times$2 pixels observations. We observe that with larger observations or longer communications, the prediction quality   further increases up to 97.75\%.
		
	}
	
	
	\section{Problem Statement} 
	
	Suppose that $N$ identical agents are in a static unknown environment. Let the agents start from a pre-determined spatial configuration. At each time step, each agent is capable of collecting a partial observation from the environment, perform{ing} some local data processing, and communicat{ing} the result with neighboring agents. The agents are allowed to communicate over a directed graph, where neighbors of an agent are those   whose messages can be received by that agent. We assume that each agent knows its own pose with respect to the environment and can take certain actions to move and update its pose at each time step. The collective objective of these agents is to classify the instance of the environment from a finite number of possibilities $\{1,2,\dots,M\}$ over a finite time horizon. 
	
	In order to decentralize the process throughout the execution, the actions by each agent should be decided solely based on the local information available to them. As a result, agents need to learn how to communicate, extract relevant features and specifications from partial observations, navigate in the environment, and reliably solve the classification problem.  
	
	In Section \ref{sec:three}, we propose a modular architecture for the network of multiple agents and discuss the details of different modules of an agent   to achieve decentralized {classification}. In Section  \ref{sec: reinforce-learn}, we demonstrate the required steps for learning the design parameters using reinforcement learning.

	

	
	
	

	\section{Connections to the Literature}
	
	In most of the multi-agent reinforcement learning literature, it is typical to assume that the agents are non-identical since their respective policies might be very distinct\cite{lowe2017multi,zhang2018fully}. Even though this assumption might be non-avoidable in some applications, assuming the existence of a common shared policy for all agents can be helpful, especially when their policies can be distinguished via some visible characteristics e.g., their location. Among the literature, \cite{khan2018collaborative} has considered a similar setup for the homogeneity of the agents.
	{This assumption  provides an advantage: the agents can learn from each others' experience. A similar idea is followed in \cite{silver2013concurrent}, where an agent interacts with multiple instances of an environment, allowing her to learn from a concurrent stream of experiences. Even though they study a single-agent scenario, the homogeneity assumption allows approaching this multi-agent problem as a single-learner problem.

	{In series of works on multi-agent reinforcement learning, the focus is on cases where the data is spread over a number of agents and the goal is to conduct the \emph{training} in a distributed or decentralized manner. For instance, a promising framework for this purpose is the Federated Learning, where the goal is to conduct the stochastic gradient descent by combining partially computed gradients over different agents \cite{mcmahan2016communication,bonawitz2017practical,bonawitz2019towards}. Contrary to this line of research, we consider settings in which the agents need to communicate throughout the \emph{execution} as well. 
		
		In a more related paper, the authors of \cite{zhang2018fully} consider a value function approximation approach for decentralized learning with interconnected agents and bring operational guarantees in the case of  linear value functions. A similar multi-agent learning problem is addressed in \cite{foerster2016learning}, where deep neural networks are used as function approximators.  A generalization of this problem within an actor-critic scenario appears in \cite{foerster2018counterfactual}. 
		Contrary to these works, we address the case where each data point is observed by a number of agents that are collaborating to fulfill a task (e.g.{,} see Fig. \ref{fig:five}), while the next set of observations are affected by  (locally) decided actions of the agents. Moreover, in our settings, the reward for reinforcement learning becomes differentiable. This necessitates revisiting the derivation of the policy gradients. }

	{Another related line of research concerns the end-to-end design of distributed or decentralized control architectures, where the goal is to learn optimal control laws in a setting with multiple dynamic agents \cite{nguyen2017selectively,fazel2018global,alemzadeh2018distributed}. }
		
		


		\subsection{Notations and Preliminaries} 
		
		The set of real numbers, nonnegative real numbers, and nonnegative integer numbers are denoted by $\R$, $\R_+$ and $\Z_+$, respectively. Other sets are denoted by script letters; e.g.{,} $\mathcal{A}$ while their cardinality is denoted by $|\mathcal{A}|$. 
		We use bold letters to denote maps; e.g.{,} $\mathbf g(x)$. The trainable parameter of a map is denoted by $\theta_i$ appearing   as a subscript; e.g.{,} $\mathbf f_{\theta_1}$.   The map $\mathbf{{SoftMax}}(x):~\R^M \rightarrow \R_{+}^M$ is standard softmax with   $k$'th element   given by
		$
		{\exp(x_k)}/{  \sum_{i=1}^M \exp(x_i) }. 
		$
		A directed graph $\G$ is characterized by a set of nodes (or vertices) $\mathcal{V}:=\{1,2,\dots,N\}$ and a set of directed edges (or arcs) denoted as 
		$
		\E \subset \{(i,j):~i,j \in \V,~i\neq j\}.
		$
		We say   node $j\in \V$ is an in-neighbor of node $i\in \V$ if $(j,i) \in \E$, and denote the set of all of its in-neighbors by $\mathcal{N}_i:=\{ j \in \V:~ (j,i) \in \E\}$.

		
		\section{Architecture of the Multi-Agent Network}\label{sec:three}
		
		We discuss details of a modular design  to solve the image classification problem using multiple autonomous agents. 
		
		

		\subsection{Temporal Evolution of Agents' Beliefs}  
		
		In order to enable learning long-term dependencies during the classification task, we equipped each agent by a dynamic module using a Long Short-Term Memory (LSTM) cell \cite{hochreiter1997long}. The role of this module is to encapsulate the aggregate belief of an agent throughout the task.  Following the widely accepted terminology \cite{goodfellow2016deep}, let us denote the hidden state and  cell state of the LSTM module on agent $i \in \mathcal{V}$ at time $t \geq 0$ by $h_i(t) \in \R^n$ and $c_i(t) \in \R^n$, respectively. Each agent updates its own belief upon { receiving new observations,} communicating with its neighbors, and forming an information input $u_i(t) \in \R^{3n}$ that contains three components: features of local observations, the average of the decoded messages received from its neighbors, and information about its location. The time evolution of the {belief} LSTM module is governed by 
		\begin{align}\label{eq:open_loop} 
		\begin{bmatrix}
		h_i(t+1)  \\ c_i(t+1)
		\end{bmatrix}=\mathbf f_{\theta_1}\left ( \begin{bmatrix}
		h_i(t)  \\ c_i(t)
		\end{bmatrix}, u_i(t) \right ), 
		\end{align}
		where  nonlinear map $\mathbf f_{\theta_1 }:~\R^{2n} \times \R^{3n} \rightarrow \R^{2n}$ is parametrized by a trainable vector $\theta_1 \in \R^{n_f}$. In the following subsections, we discuss each component of the information input   to the LSTM module.  
		
		\subsection{Agent Motion and Stochastic Action Policy}
		
		Let us represent  the spatial state (or pose) of agent $i \in \V$ by $p_i(t) \in \R^d$ and the finite set of all possible actions, which agents can take,  by $\mathcal{A}$. Each agent moves in the spatial domain according to dynamics 
		\begin{align}\label{eq:pmap}
		p_i(t+1)=\mathbf g \big(p_i(t),a_i(t+1)\big),
		\end{align}    
		where $\mathbf g:\R^{d} \times \mathcal{A} \rightarrow \R^d$ is a { known} transition map and action $a_i(t+1)$ is  sampled  from set $\mathcal{A}$ according to a probability mass function  $\pi: \mathcal{A}  \rightarrow \R$} that is computed as follows. We use a state-dependent stochastic action policy by updating the action probabilities according to 
		\begin{align}
		\pi(a)={\boldsymbol \pi}_{\theta_3}(a, \hat h_i(t+1)),\label{action-policy}
		\end{align}
		where  $a$ is an action in $\mathcal A$ and $\hat h_i(t+1)$ is the hidden state of {the decision} LSTM unit whose dynamics are governed by 
		\begin{align}\label{eq:ulstm}
		\begin{bmatrix}
		\hat h_i(t+1)  \\ \hat c_i(t+1)
		\end{bmatrix}=\mathbf f_{\theta_2}\left ( \begin{bmatrix}
		\hat h_i(t)  \\ \hat c_i(t)
		\end{bmatrix}, u_i(t) \right ). 
		\end{align} 
		This LSTM unit is fed with exactly the same information input $u_i(t)$ as the belief LSTM module \eqref{eq:open_loop}. 
		
	  We consider one fully connected layer with a ReLU {activation} {followed by another} fully connected linear layer {to represent the map $\pi$, where we denote by $\theta_3 \in \R^{n_\pi}$ the corresponding trainable parameters}.

		\begin{example}\label{ex:one} For a flying robot that can translate and rotate in the $3$D space,  a natural choice for  spatial state $p_i(t)$ is a vector in $\R^6 $  created by stacking three  position components of the robot and three  Euler angles describing its orientation (relative to the environment {frame}). 
		\end{example}

		
		\subsection{Inter-Agent Communication Architecture}
		
		The agents are allowed to communicate over a  directed graph $\G$ with  node set $\V$ and {arc}  set $\E$. For distinct agents $i,j \in \V$, $(i,j)\in \E$ implies that agent $j$ receives messages from agent $i$. Each agent generates a  message\footnote{It is assumed that when agent $i$   broadcasts its message $m_i(t) \in  \R^{n_m}$ at time $t$, all neighboring agents receive identical copies of that message.} using its belief hidden state according to 
		\begin{align}\label{eq:messages}
		m_i(t)=\mathbf m_{\theta_4}(h_i(t)),
		\end{align}
		where map $\mathbf m_{\theta_4}:\R^n \rightarrow \R^{n_m}$ is parameterized by a trainable vector $\theta_4 \in \R^{n_e}$.  A sequence of two layers   is considered for this map: a fully-connected   layer  with   ReLU activation followed by a fully connected linear layer for the output.

		
		\subsection{Observation Model and Feature Extraction}\label{subsec:obs-model}
		
		Suppose that agent $i$ at time $t$ collects (partial) observation $o_i(t) \in \R^{f \times f}$.     It is assumed that agents' observations can be completely characterized by its pose $p_i(t)$. Thus, 
		\begin{align}\label{eq:omap}
		o_i(t)=\mathbf o(I,p_i(t)),
		\end{align}
		where $I\in \R^{n_I \times n_I}$ is the entire  image. This identity can be interpreted as the \emph{repeatability} property of the observations: 
		two agents with different past history will observe the same image provided that they both have identical poses at the observation time. The relevant features  of an observation can be extracted by a {parameterized}  map
		\begin{align}
		b_i(t)=\mathbf b_{\theta_5}(o_i(t)),
		\end{align}
		where ${\theta_5} \in \R^{n_c}$ is a trainable vector. The nonlinear map $\mathbf b_{\theta_5}:~\R^{f \times f } \rightarrow \R^n$ results from the following three layers: two single layer  {{convolutional neural networks}}   followed by { vectorization and} a fully connected layer.
		
		\begin{example} Fig. \ref{fig:five},   illustrates a case in which the spatial state is simply the location of the agent (relative to the image) and observation map \eqref{eq:omap} crops a subset of the image based on its position. Moreover,  the map describing the motion of the robots, according to \eqref{eq:pmap}, has resulted in   horizontal and vertical translations of the agents across the image.  
		\end{example}
		
		\begin{example} \label{ex:two} Let us consider the settings of Example \ref{ex:one}, where a  camera is mounted on the robot. Then, map $\mathbf o(.)$  in \eqref{eq:omap} for this case is the {projection map of the camera. For a camera, this map}  is completely characterized by the position and orientation of the robot with respect to the environment {(i.e., camera extrinsics)}.  
			
			
		\end{example}
		
		\subsection{Structure of Information  Inputs} 
		
		In the previous subsections, we explained the details of {the} belief dynamics, agent motion and actuation, observation processing, communication, and decision-making modules on each agent. The same information input is fed to both LSTM modules in \eqref{eq:open_loop} and \eqref{eq:ulstm}. We design the information input as a vector in $\R^{3n}$ with components  
		\begin{align}\label{eq:input_shape}
		u_i(t)=\begin{bmatrix} b_i(t)^T  
		& \bar d_i(t)^T 
		& \lambda_i(t)^T
		\end{bmatrix}^T.
		\end{align}
		All these three components can be calculated using locally accessible data {as we elaborate below}. In Subsection \ref{subsec:obs-model}, it was shown that $b_i(t)$ contains {the} features of the (partial) observation.   
		
		After communicating with neighbors, each agent 
		decodes the received messages using a   map  $\mathbf d_{\theta_6}:\R^{n_m} \rightarrow \R^n$ to get 
		\begin{align}
		d_i(t)=\mathbf d_{\theta_6} (m_i(t)), 
		\end{align}
		where $\theta_6$ is a trainable vector. We consider a fully connected layer with a ReLU activation for this map.  Then, each agent takes the average of the received messages   to find 
		\begin{align}
		\bar d_i(t)=\dfrac{1}{\Delta_i} \sum_{(j,i) \in \mathcal{E}} d_j(t),
		\end{align}
		in which $\Delta_i$ is the in-degree of node $i$ in graph $\mathcal{G}$. This, $\bar d_i(t)$ is the aggregate message received by agent $i$ at time $t$. 
		
		It is useful for agents to tag their beliefs and information by their spatial state. This can be done by  the following map  
		\begin{align}
		\lambda_i(t)=\boldsymbol \lambda_{\theta_7}(p_i(t)){ ,}
		\end{align}
		where $\boldsymbol \lambda_{\theta_7}:\R^d \rightarrow \R^n$ is a parametrized map with a trainable vector $\theta_7$. 
		
		In the final step, we close the loop by applying information input \eqref{eq:input_shape} to \eqref{eq:open_loop} and \eqref{eq:ulstm}.
		
		
		\section{Decentralized Prediction and Classification}
		
		Recall that the image should be classified from $M$ categories, while we have $T$ rounds of observation and communication.  To do this, first the raw prediction vector by agent $i$ is evaluated using the final cell state and a map  $\mathbf q_{\theta_8}:~\R^n \rightarrow \R^M$ as  
		\begin{align}
		q_i=\mathbf q_{\theta_8}(c_i(T)).
		\end{align}
		We use a fully-connected linear layer with ReLU activation   {followed by a fully connected linear layer in place of this map}. Then, we run a distributed average consensus algorithm over a strongly connected directed graph. Upon this averaging, on each agent, we will have   {a shared} prediction vector $\bar q \in \R^M$, which is given by 
		\begin{align}\label{eq:qbar}
		\bar q=  \dfrac{1}{N}\sum_{i=1}^N q_i. 
		\end{align}
		It has been shown that if the communication graph is strongly connected, this task can be conducted in a completely decentralized manner \cite{dominguez2011distributed}.  Finally, {each agent evaluates the system-wide}   prediction category using
		\begin{align}
		q_c= \underset{ j \in \{1,\dots,M\}}{\mathrm{argmax}} ~~ \mathbf{{SoftMax}}(\bar q).
		\end{align}
		
		One should note that in the current approach, we do not force the agents to reach a consensus on the predicted category, but rather we {combine} the beliefs due to sequences of partial observations to produce a single prediction.    
		
		\begin{figure}[t]
			\centering
			\includegraphics*[width=7.607cm]{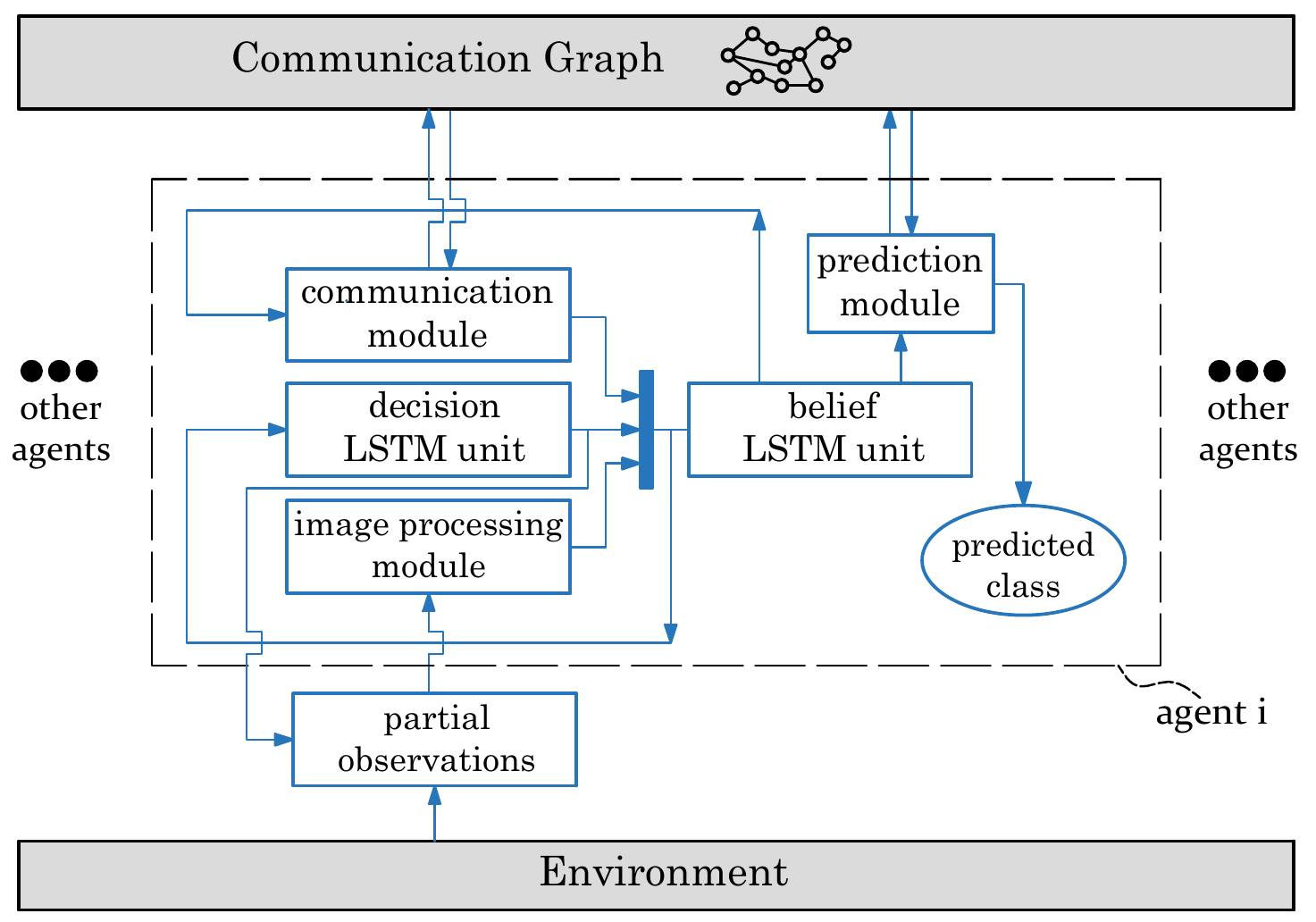}
			\caption{The diagram illustrating the essence of our framework. 
			}
			\label{fig:diagram}
		\end{figure}
		
		

		In Fig.   {\ref{fig:diagram}} we have illustrated the {information flow} of the framework that has been described throughout the section. These settings and steps can be summarized to build Algorithm \ref{alg:MARL}, whose output  is   prediction category $q_c\in \{1,\dots,M\}$ (shared by all agents).

		
		\begin{algorithm}[t]
			\caption{Multi-Agent Classification (Execution)}
			\label{alg:MARL}
			\begin{algorithmic}  
				\algnewcommand\algorithmicinitz{\textbf{initialize:}}
				\algnewcommand\Init{\item[\algorithmicinitz]}
				\renewcommand{\algorithmicrequire}{\textbf{input:}}
				\renewcommand{\algorithmicensure}{\textbf{output:}}
				\Require  Input image $I \in \R^{n_I \times n_I }$ \\ ~~~~~ initial spatial states $p_1(0),\dots,p_N(0)$
				\Ensure  prediction category $q_c$
				\vspace{0.15cm}
				\Init  \For {$i \in \mathcal V$}
				\State initialize the states $h_i(t) \leftarrow 0$, $ c_i(t) \leftarrow 0$
				\For {$j \in \mathcal{N}_i$}
				\State initialize the messages $ m_j(0) \leftarrow 0$
				\EndFor
				\EndFor  \vspace{1mm}
				\For {$t=0$ to $T-1$}
				\Comment communication \& observation  
				\For {$i \in \mathcal V$}
				\State conduct the observation $o_i(t)\leftarrow \mathbf o(I,p_i(t))$
				\State map the observation  $b_i(t)\leftarrow \mathbf b_{\theta_5}(o_i(t))$
				\For {$j \in \mathcal{N}_i$}
				\State decode       message  $d_j(t) \leftarrow \mathbf d_{\theta_6} (m_j(t)). $
				\EndFor    
				\State find average   message 
				$
				\bar d_i(t)\leftarrow \dfrac{1}{\Delta_i} \displaystyle \sum_{(j,i) \in \mathcal{E}} d_j(t)
				$ \vspace{1mm}
				\State map the spatial state $\lambda_i(t)\leftarrow \boldsymbol{\lambda} _{\theta_7}(p_i(t))$ \vspace{1mm}
				\State form   input  $u_i(t)\leftarrow 
				\begin{bmatrix}
				b_i(t)^T  
				& \bar d_i(t)^T 
				& \lambda_i(t)^T
				\end{bmatrix}^T.$
				\State  run the belief LSTM unit  \eqref{eq:open_loop} 
				\State evaluate   message $m_i(t+1)\leftarrow \mathbf m_{\theta_4}(h_i(t+1))$
				
				\State  run the decision LSTM unit  \eqref{eq:ulstm} 
				
				\State update   policy   distribution $ \boldsymbol \pi_{\theta_3}(.\,|\hat h_i(t+1))$
				\State samples action $a_i(t+1)$ based on $ \pi$
				\State update   spatial state $p_i(t+1)\leftarrow \mathbf g(p_i(t),a_i(t+1))$
				\EndFor 
				\vspace{0.05cm}
				%
				\vspace{0.05cm}
				\EndFor  \vspace{0.02 mm} 
				\For{$i \in \V$} \Comment local raw predictions
				\State find raw prediction vector $q_i\leftarrow \mathbf q_{\theta_8}(c_i(T))$
				\EndFor
				\State conduct the distributed average consensus $\bar q \leftarrow  \dfrac{1}{N}\displaystyle \sum_{i=1}^N q_i $
				\State find the prediction category $$q_c \leftarrow \mathbf{{SoftMax}}\big (\underset{i \in \{1,\dots,M\}}{\mathrm{argmax}} ~~\bar q \big ) $$

			\end{algorithmic}
		\end{algorithm}

		\section{Reinforcement Learning}\label{sec: reinforce-learn}
		
		{We derive a generalization of the vanilla policy gradient algorithm, which utilizes the intrinsic differentibilty of the rewards for simultaneous training of both prediction and motion planning parameters.}
		First, let us stack our parameters as a  single design parameter according to 
		\begin{align}
		\Theta:=\left [\theta_1^T,\theta_2^T,\dots,\theta_8^T \right ]^T. 
		\end{align} 
		Next, let us denote all  trajectories with positive probability of occurrence by  $\mathcal{T}$. Suppose that in a sample execution $\tau \in \mathcal{T}$, image $I$ corresponds to category $j \in \{1,\dots,M\}$ (i.e., its actual category is $j$). Then, we define the reward corresponding to the outcome of this sample trajectory 
		\begin{align}
		r_\tau:=-\mathbf f_l(\bar q_\tau-e_j),
		\end{align}
		where $\mathbf f_l$ is a differentiable  nonnegative loss function {(e.g.$L_2$ Norm)}, $\bar q_\tau$ is the prediction at the end of this sampled trajectory, and  $e_j \in \R^M$ is the unit coordinate vector in direction $j$.  
		Based on the goal of this problem, we define our objective function as 
		\begin{align}
		J(\Theta) =\mathbb{E} \{r_\tau\}  ={\sum_{\tau \in \mathcal{T}}  P_\tau \,  r_\tau }. 
		\end{align}
		{Here $P_\tau$ is the probability of sampling trajectory $\tau$ for a given $\Theta$. Therefore}, we need to solve the optimization problem
		\begin{align}
		\underset{\Theta}{\mathrm{maximize}}{}~~ J(\Theta). 
		\end{align}
		The gradient of $J$ with respect to $\Theta$ can be written as 
		\begin{align}
		\nabla_{\Theta} J\,=\,\sum_{\tau \in \mathcal{T}} r_\tau \nabla_{\Theta}  P_\tau +  P_\tau \nabla_{\Theta}  r_\tau.
		\end{align}
		{Let us drop   index $\Theta$ for simplicity.} Using the well-known {gradient derivation} technique { similar to that of} the REINFORCE algorithm \cite{sutton2000policy}, we can write 
		\begin{align}\label{eq:nabla}
		\nabla J &\,=\,\sum_{\tau \in \mathcal{T}} P_\tau \nabla (\log P_\tau)r_\tau + P_\tau \nabla r_\tau \\
		& \notag \, =\, \mathbb{E}\{\nabla (\log P_\tau)r_\tau+\nabla r_\tau\}.
		\end{align}
		Let us  execute Algorithm \ref{alg:MARL} for $N_{r}$ independent experiments. Then, for each sample $k=1,\dots,N_r$, we use  
		$p^{(k)}$ to denote the probability that this particular trajectory is selected. Now, inspired by \eqref{eq:nabla}, we define the proxy sampler for $J$ to be 
		\begin{align}\label{eq:hatJ}
		\hat J\,:=\,\dfrac{1}{N_r} \Big ( \sum_{k=1}^{N_r} \log p^{(k)} r_d^{(k)}+r^{(k)} \Big ),
		\end{align}
		where quantity $r_d^{(k)}$ has a value equal to $r^{(k)}$, but has been detached from the gradients. {This means that a machine learning framework should treat $r_d^{(k)}$ as a non-differentiable scalar during training.  } 
		Then, we inspect that 
		\begin{align}
		\mathbb{E} \left \{\nabla \hat J \right \}\, =\, \nabla J,
		\end{align}
		i.e., $\nabla \hat J$ is an unbiased estimator of $\nabla (\log P_\tau)r_\tau+\nabla r_\tau$ that appears in \eqref{eq:nabla}. 
		Therefore, it is justified to follow the approximation for the gradient given by 
		\begin{align}\label{eq:firstapprox}
		\nabla J \, \approx \, \nabla \hat J. 
		\end{align} 
		Note that the first term in summation \eqref{eq:hatJ} is identical to the quantity that is derived in the policy gradient method with a reward that is independent of the parameters (i.e., identical to REINFORCE algorithm).  The second term accounts for the fact that the reward in our settings directly depends on parameter $\Theta$:  for two {different set of parameters}, if the agents receive exactly the same sequences of observations and take exactly the same actions, still the reward explicitly depends on the parameters of the network {(e.g.{,} weights of the convolution layers or fully connected layers)}.  
		

		\section{Numerical Experiments}
		

		\begin{figure}[t]
			\centering
			\fbox{\includegraphics*[width=4.0cm]{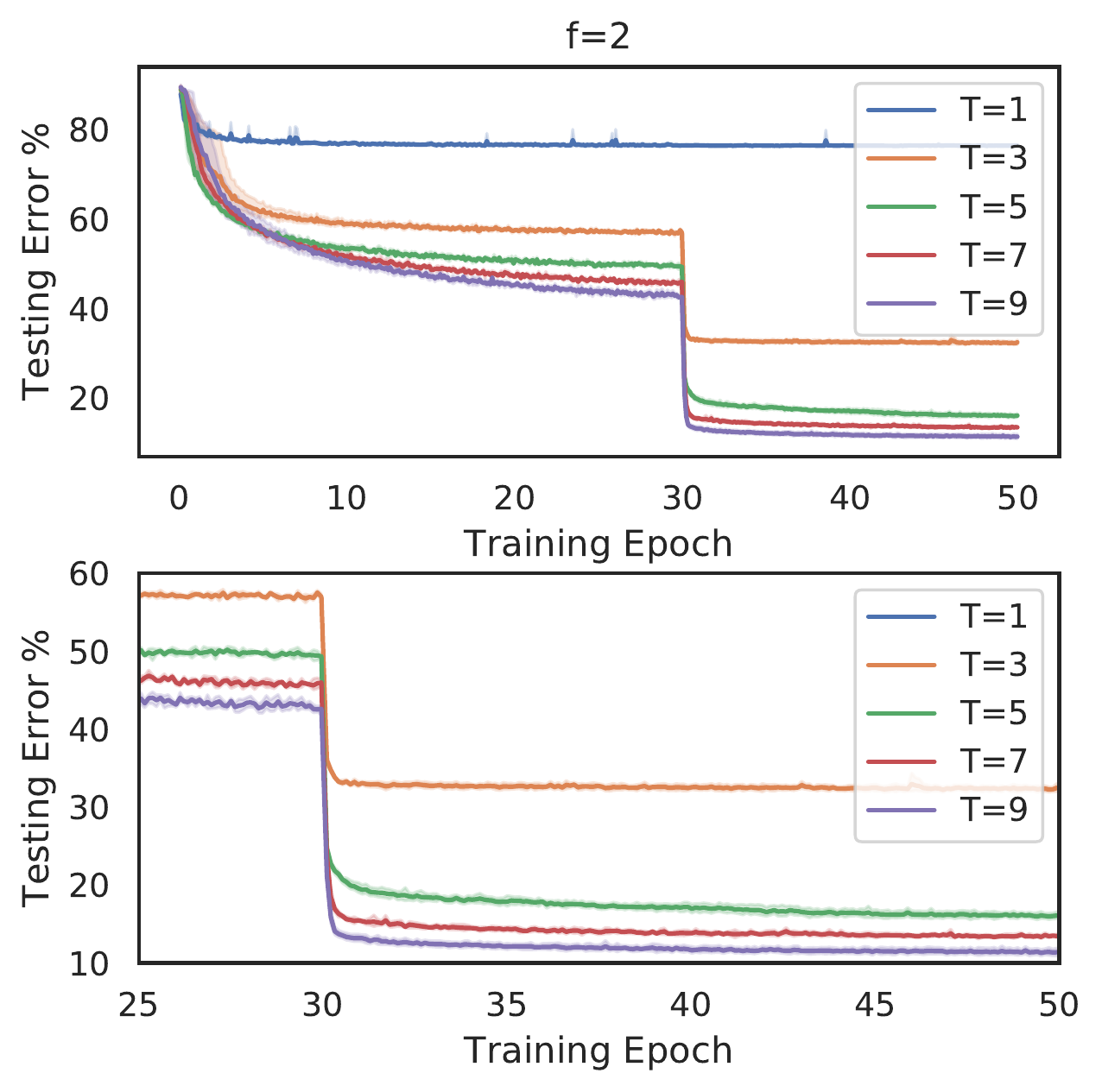}
				\includegraphics*[width=4.0cm]{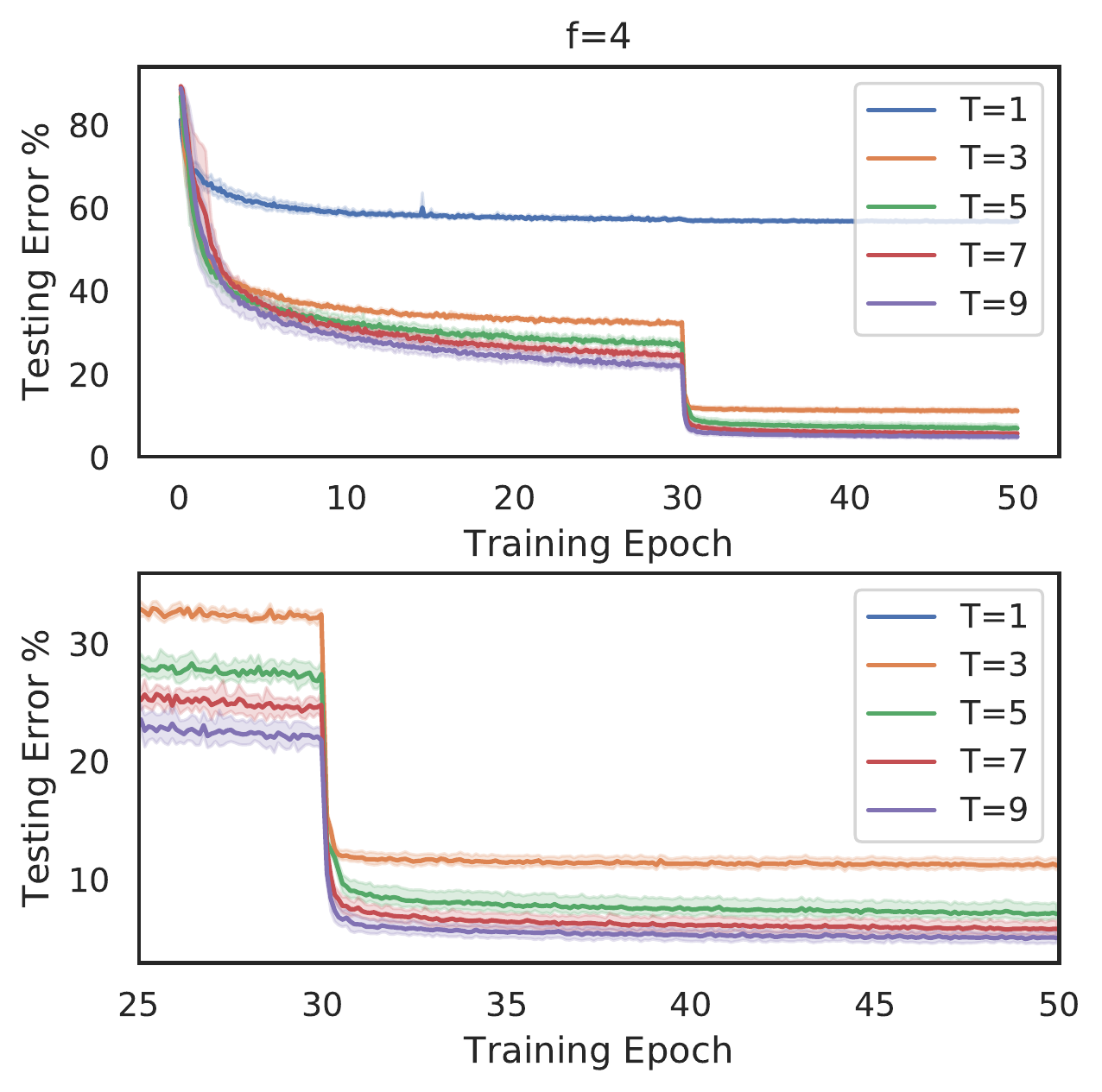}}
			\\ 
			\fbox{\includegraphics*[width=4.0cm]{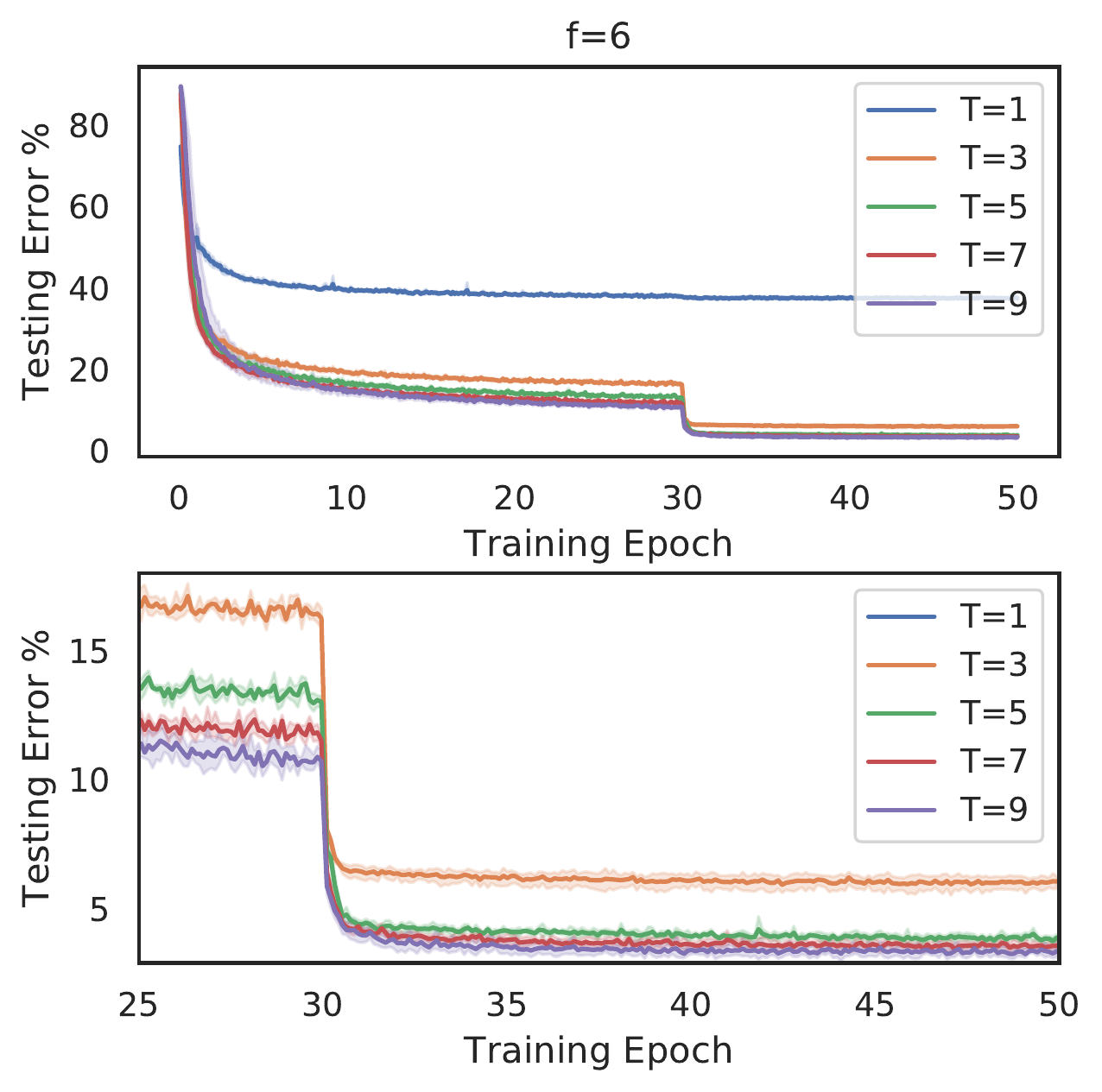}
				\includegraphics*[width=4.0cm]{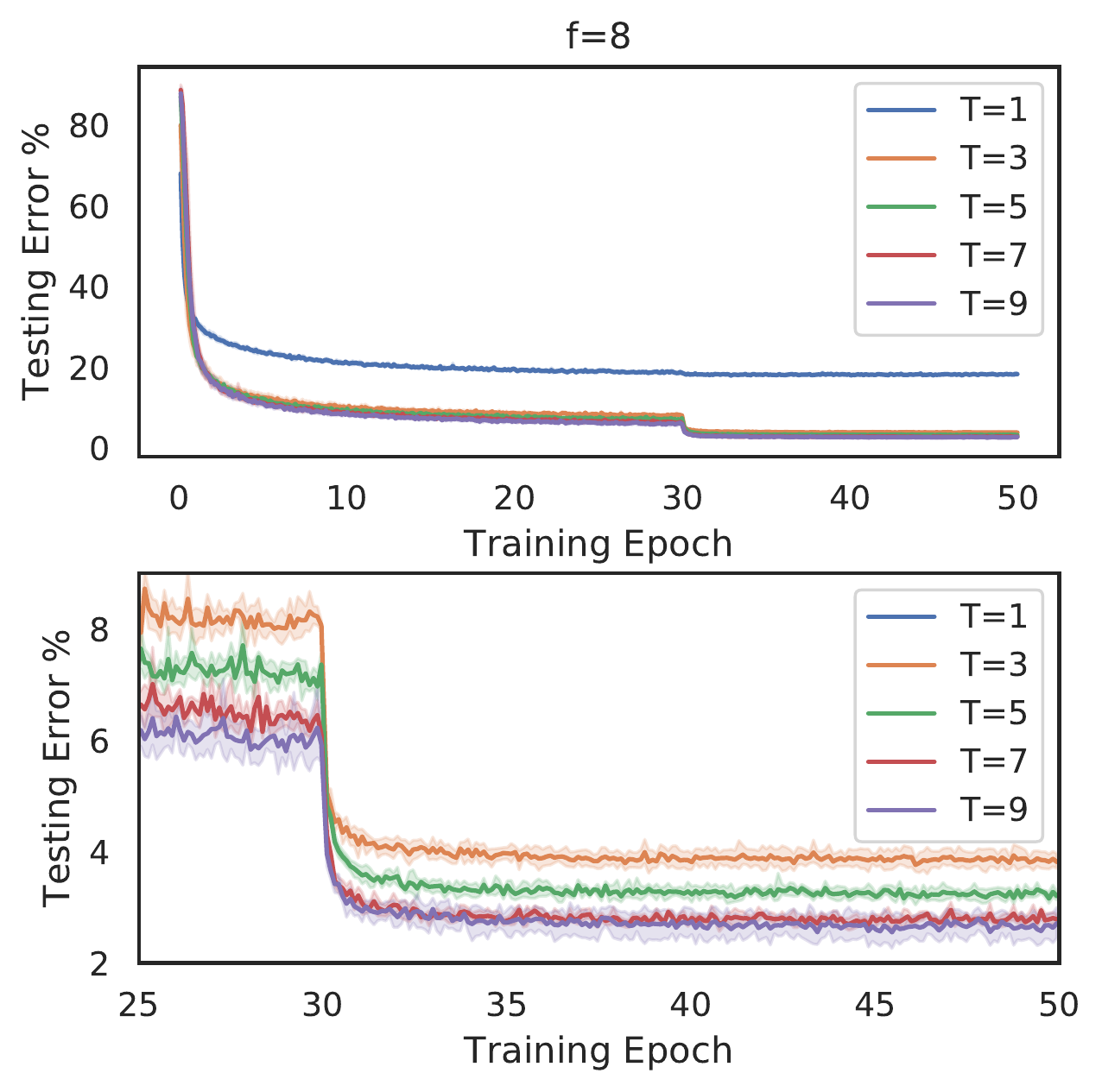}}
			\caption{The testing accuracy for different frame sizes $f$ and time horizons $T$ versus the number of training epochs. }
			\label{fig:progress}
		\end{figure}
		
		\begin{figure}[t]
			\centering
			\includegraphics*[width=7.0cm]{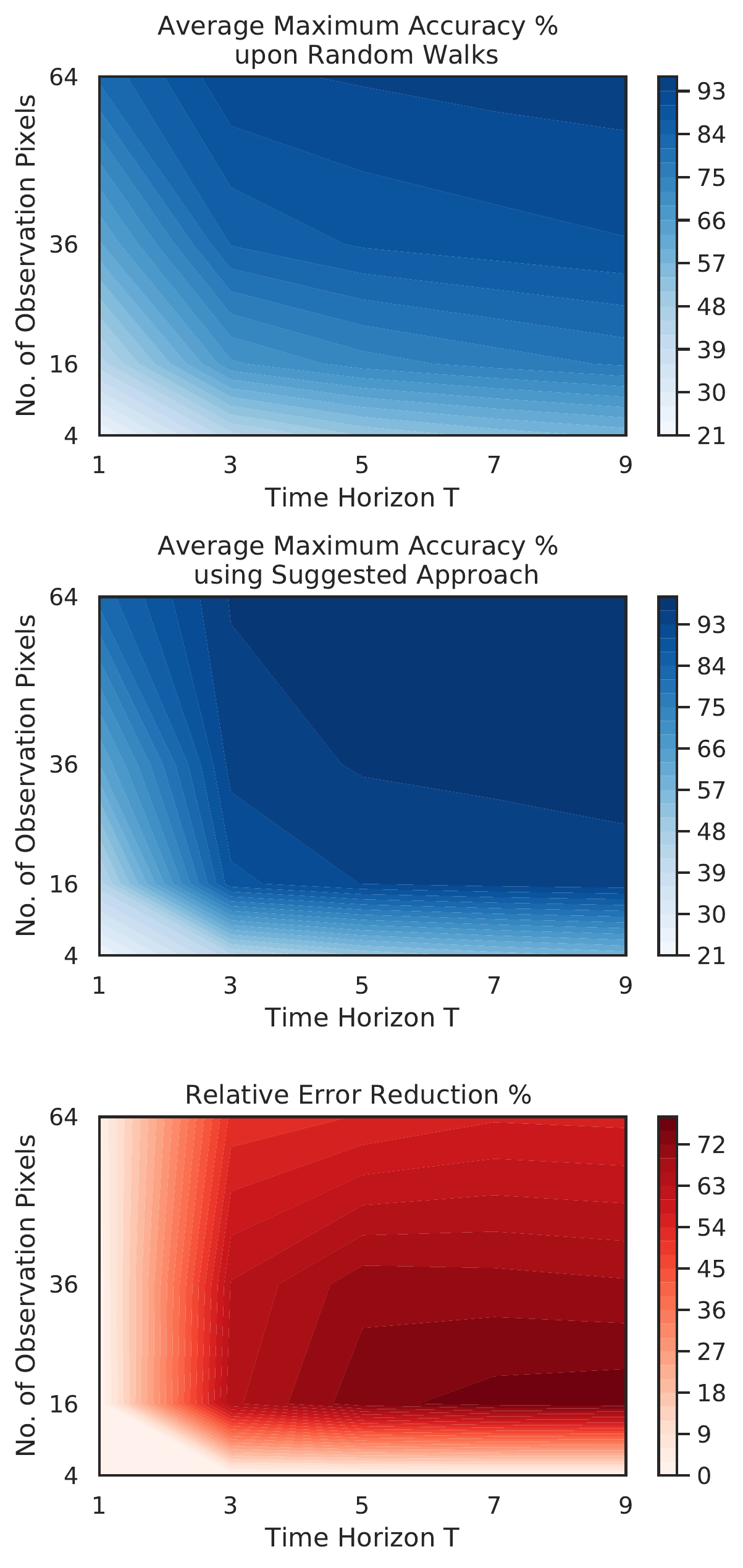}
			\caption{Average maximum testing accuracy versus frame size $f$ and time horizon $T$ after $30$ epochs with random walks (top figure){.}  {In the middle one, we trained extra $20$ training epochs for the motion planning policy. } {T}he last figure   illustrate{s} the error reduction as a result of {the} design of coordination policy instead of random walks. }
			\label{fig:heatmap}
		\end{figure}

		{We use the MNIST {dataset} of handwritten digits \cite{lecun1998gradient} to test the proposed   learning algorithm. The dataset consists of 60,000 training images and 10,000 testing images, where each image {has} $28 \times 28$ pixels. We suppose that each agent {may} observe a portion of the image  that   {has} $f \times f$ pixels. The spatial variable $p_i$ is the pixel coordinate of the top left corner of the observation window. The possible movements by each agent can be characterized by 
			\begin{align}\label{eq:ourA}
			\mathcal{A}=\{\text{up},\text{down},\text{left},\text{right}\}.
			\end{align}
			By each movement, the agent is translated {in} the desired direction by $f_m$ pixels. 
			If the sampled action is infeasible, then the agent {remains} at its location at that time instant. In Fig. \ref{fig:five}, as an example, we have illustrated an image {from these data} and {the observations that three agents receive during the horizon.} { In all experiments of this section,} we choose a mini-batches to have $64$ images {during the} training. We also choose the {variable size} of LSTM unit to be $n=64$, {the number of neurons of all fully connected layers to be $64$,} and the dimension of the broadcasted messages to be $n_m=12$. We have implemented this approach using the machine learning {framework} PyTorch \cite{paszke2017automatic}.}

		%
		%
		%

		\noindent{\textbf{Testing Accuracy Results:}} We consider $N=2$ agents that are communicating over {the only option} for a strongly connected graph; i.e., graph with arc set $\E=\{(1,2),(2,1)\}$. We choose $N_r=30$   and conduct a parametric study by varying   observation frame size $f$ and time horizon $T$  according to $f \in \{2,4,\dots,8\}$ and  $T \in \{1,3,\dots,9\}$, respectively. For each pair of $f$ and $T$, we {train the model} for $50$ epochs. However, we break down the training  into two {stages}: first, we consider random walks for $30$ epochs. Then, we fix all of the parameters except for the decision-making module and train the model for another $20$ epochs. 
		In Fig. \ref{fig:progress}, we demonstrate the progress of the testing error versus the number of training epochs.  Also, in Fig. \ref{fig:heatmap}, we show the average maximum testing accuracy   for each pair of frame size $f$ and time horizon $T$ in the case of random walks (i.e., at the end of $30$ epochs) and also with the designed law for the movements of the agents (i.e., after additional 20 epochs of training {for motion planning}). The results suggest that  {following} a policy that governs the {actions} of the agents may significantly decrease the testing error; {for instance,} Fig. 5 implies that for $f=4$ (i.e., 16 observation pixels) and   $T=7$ communication and observation steps, the testing error has decreased by more than $70 \%$.
		
		\begin{remark}
			{The intuition behind  our two-stage method of training is that we initially train the perception,  communication, and prediction modules while agents are {learning to explore}  the environment. Once these modules are sufficiently trained, we let the agents learn how to traverse the image. The numerical experiments suggest that {this}   training {strategy} generally results in smaller testing errors{, compared to}  the case in which we simultaneously optimize the parameters of all modules. One possible justification {for} this observation is that the two-stage training   is less prone to getting stuck in local non-stationary solutions due to higher exploration in the first phase. }
		\end{remark}

		\begin{figure}[t]
			\centering
			\includegraphics*[width=7.6cm]{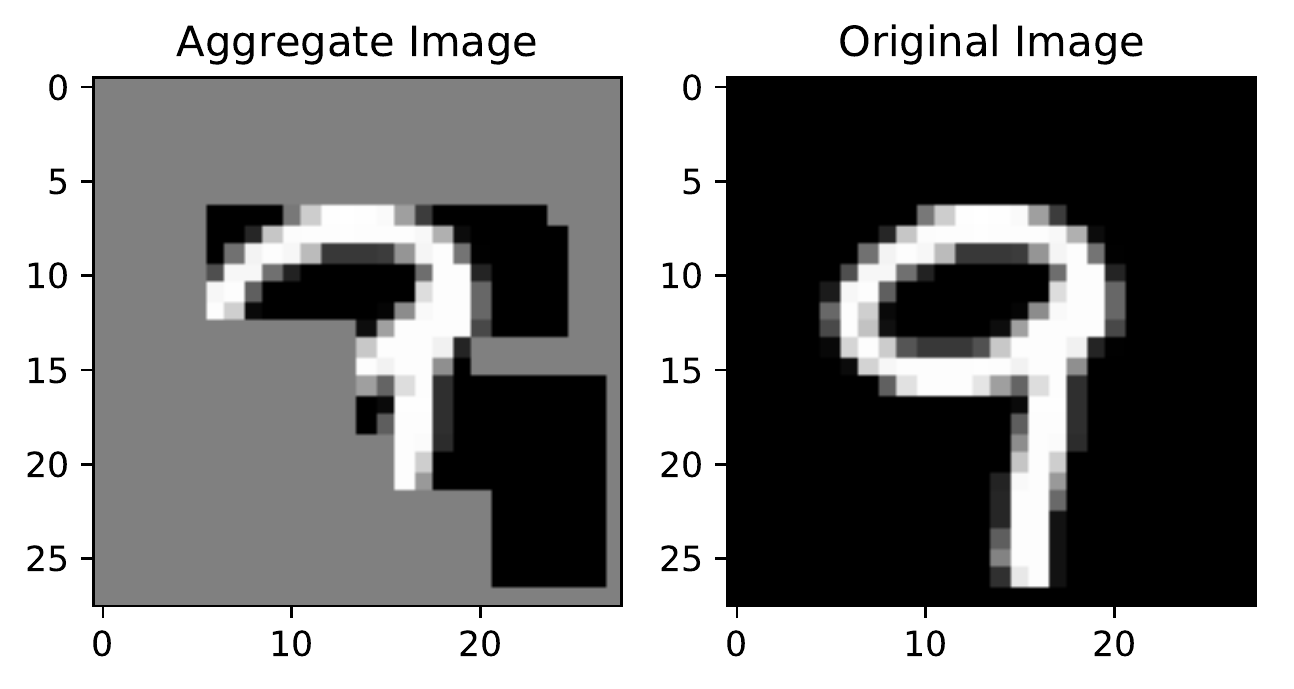}
			\caption{A sample of masked images created by putting together the observations by $3$ different agents for a time horizon of $T=3$ with $f=8$. This image is the input to the centralized image classifier  as an alternative classification approach (method \emph{(i)}). The (random) uncovered parts have been reached due to random walks.  }
			\label{fig:uncover}
			
		\end{figure}
		
		\begin{figure}[t]
			\centering
			\includegraphics*[width=8.0cm]{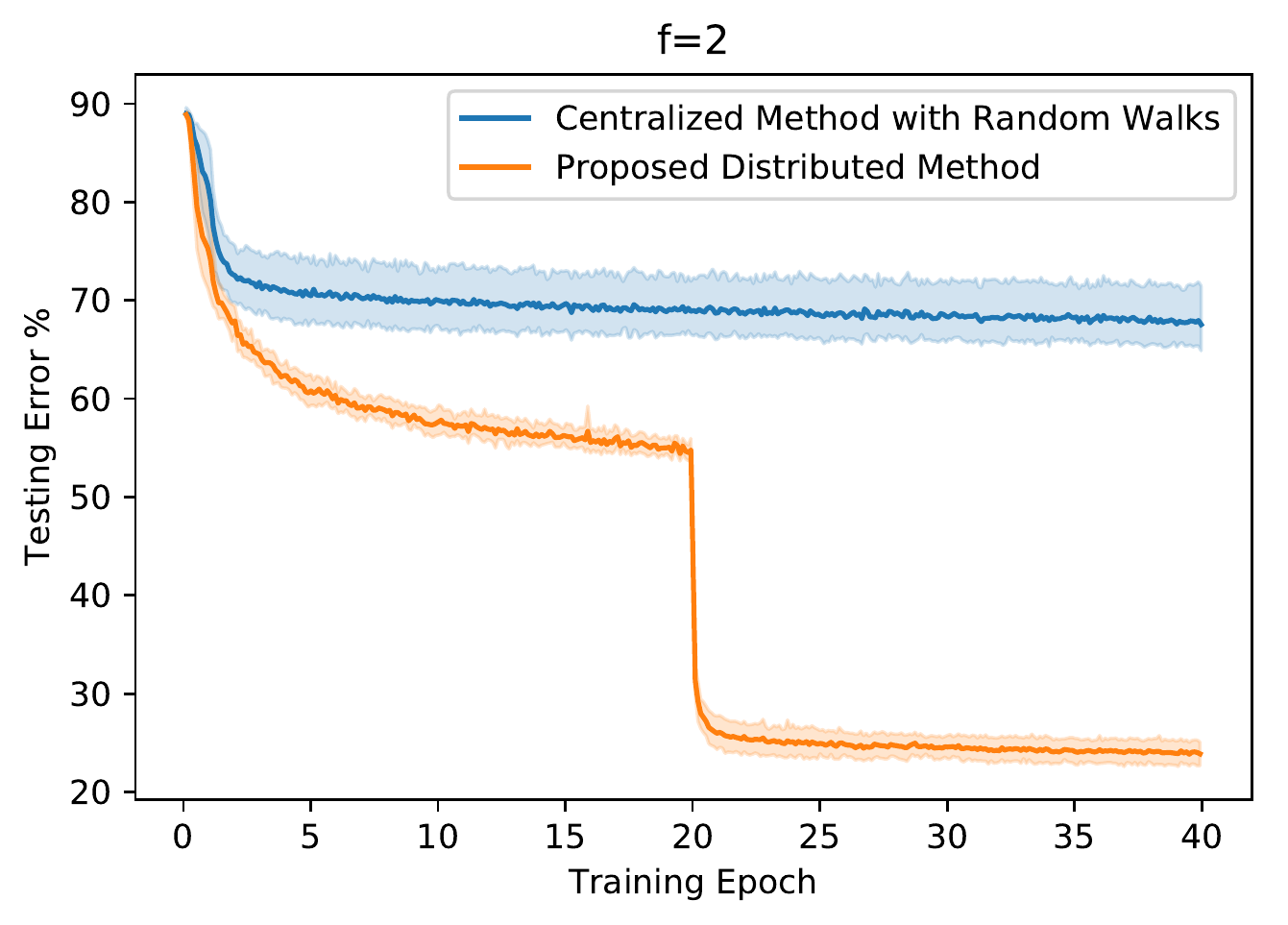}
			\caption{Comparison of the testing error when using  {a} centralized method with the suggest{ed} approach. The first $20$ epochs of training corresponds to the case for random walks, while in the next $20$ epochs we optimize the movements of the agents. }
			\label{fig:comparison}
		\end{figure}

		\begin{figure}[t]
			\centering
			\includegraphics*[width=8cm]{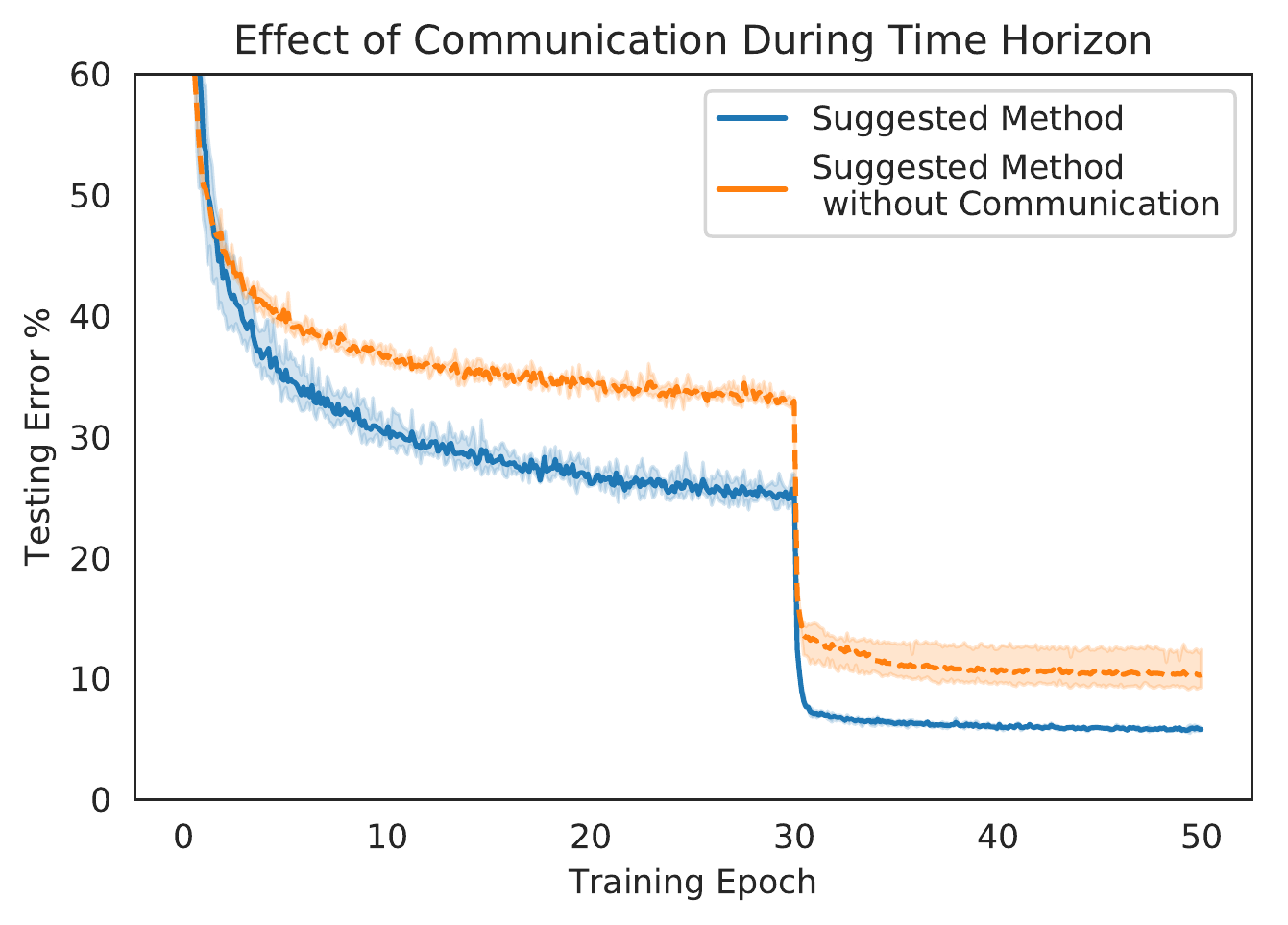}
			\caption{The result of training   with and without communications. }
			\label{fig:shareimage}
		\end{figure}
		
		\begin{figure}[t]
			\centering
			\includegraphics*[width=6.8cm]{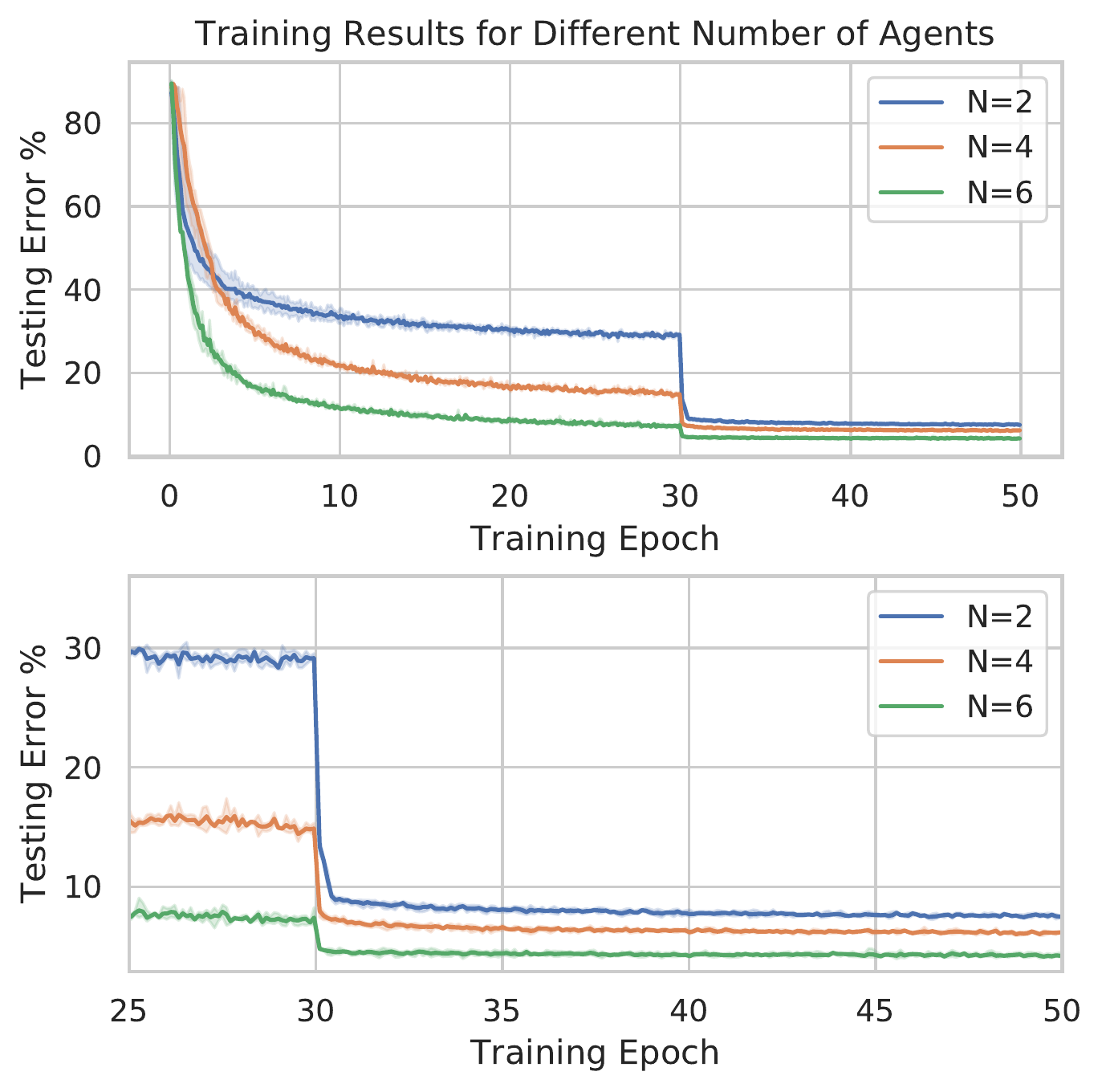}
			\caption{The training results for different number of agents. }
			\label{fig:agents}
		\end{figure}
		
		\noindent{\textbf{Alternative Classification Schemes:}} We consider two alternative methods to classify the images based on similar observations, which will be compared against our approach. In both cases, each agent independently conducts a \emph{random walk}. {\textit{(i) Centralized Classification with Random Walks:}} we collect all the images from all agents based on random movements that have equal probability in each of four directions (i.e. $1/4$). Then, we feed the resulting unmasked image to a single  {CNN} which is embedded into a  prediction vector $q \in \R^M$ (similar to $\bar q$  in \eqref{eq:qbar}). In Fig. \ref{fig:uncover}, we demonstrate a typical input of this simple centralized classification, which has the same dimensions as the images in MNIST (i.e., $f=28$). {\textit{(ii) Distributed Classification with Random Walks:}} We also consider a variant of Algorithm \ref{alg:MARL} in which instead of learning the optimal distributions for the policy,   all possible {actions} in \eqref{eq:ourA} have {an} equal probability of $1/4$.  We set the parameters to $f=2$, $N=2$ agents, $N_r=10$ samples, time horizon $T=4$, and a complete communication graph. We conduct the training with random walks for $20$ epochs, which is followed by training of the {decision-making} { motion planning} module for another $20$ epochs. We  {train the }centralized classification for $40$ epochs as well. 
		In Fig. \ref{fig:comparison}, we illustrate the results, which show that the quality of prediction using the proposed method is superior {to both cases}. 
		
\begin{remark}
			The main reason for which the performance of the centralized method is not better than our approach is that the masks   created by random motion of the agents are not optimal. 
		\end{remark}
		


		\noindent{\textbf{Effect of Communication:}} We compare the result of training for an alternative structure in which the agents do not communicate during the horizon (although, they finally do so conduct the prediction). We set the parameters to be 
		$N=2$, $f=4$, $N_r=40$, and $T=6$. As shown in Fig. \ref{fig:shareimage},   it turns out that 
		smaller testing errors compared to the case that the agents do not communicate. 
		
		\noindent{\textbf{Effect of Number of Agents:}} We consider the set of parameters $f=4$, $N_r=25$, and $T=4$ and conduct the training for a different number of agents communicating over a complete directed graph. In Fig. \ref{fig:agents}, we illustrate the result of training for $30$ epochs with random walks followed by learning the moving policies for $20$ epochs. {As expected, we observe that increasing the number of agents significantly reduces the testing error.}
		
		\noindent{\textbf{Visualization of Communicated Messages:}}
		We explore the patterns in the messages that are broadcasted by the agents using the learned communication medium.  The goal is to visualize how the agents express the shared memory within an episode for solving the task. After training, we simulated 500  sample trajectories with  $T=9$ and recorded the messages of all agents at every $t=0,\dots,8$. Then, we used the dimensionality reduction technique called t-SNE \cite{maaten2008visualizing} to produce meaningful visualizations of the messages. In Fig. \ref{fig:}, we illustrate two t-SNE plots for the messages at $t=0$ and $t=6$, which correspond to the messages before any communication and after a few rounds of communication and observation, respectively.
		In these figures, every message, {which} is initially in $\R^{12}$, is reduced to a vector in $\R^2$ and is illustrated with a color corresponding to the true label of the image.  The result of clustering at $t=0$ implies that initially, there is no meaningful pattern in the distribution of the labels. However, as the agents move across the image and communicate, they construct an internal belief about the true category. The second figure shows a t-SNE plot after 6 time-steps, which suggests that agents' beliefs are reflected in the communicated messages. In fact, we observe that the digits are now clearly clustered in their own groups; i.e., they have learned to broadcast their beliefs about the true labels to their neighboring agents.

		\begin{figure}[t]
			\centering
			\fbox{\includegraphics*[width=7.5cm]{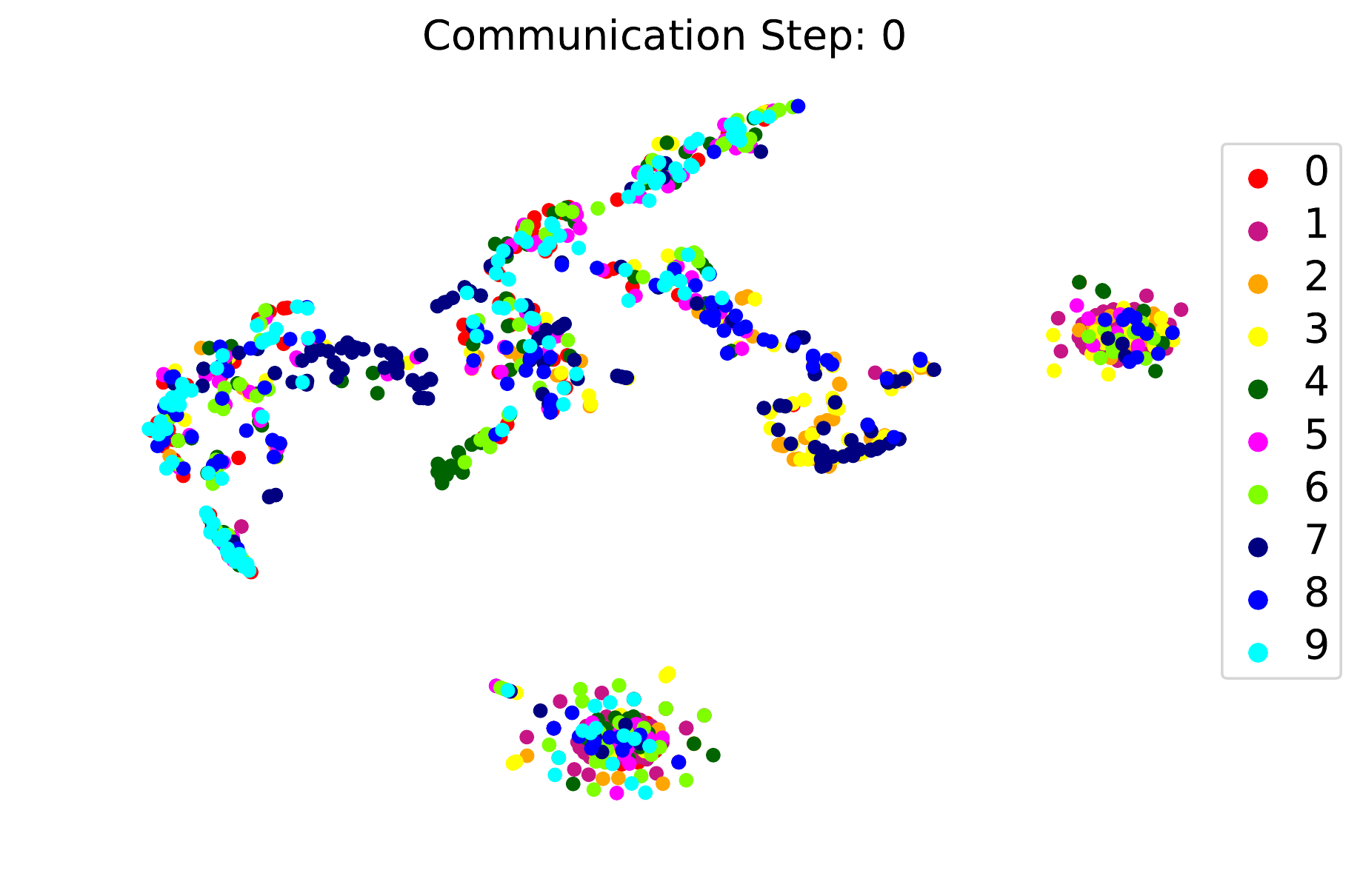}}
			\fbox{\includegraphics*[width=7.5cm]{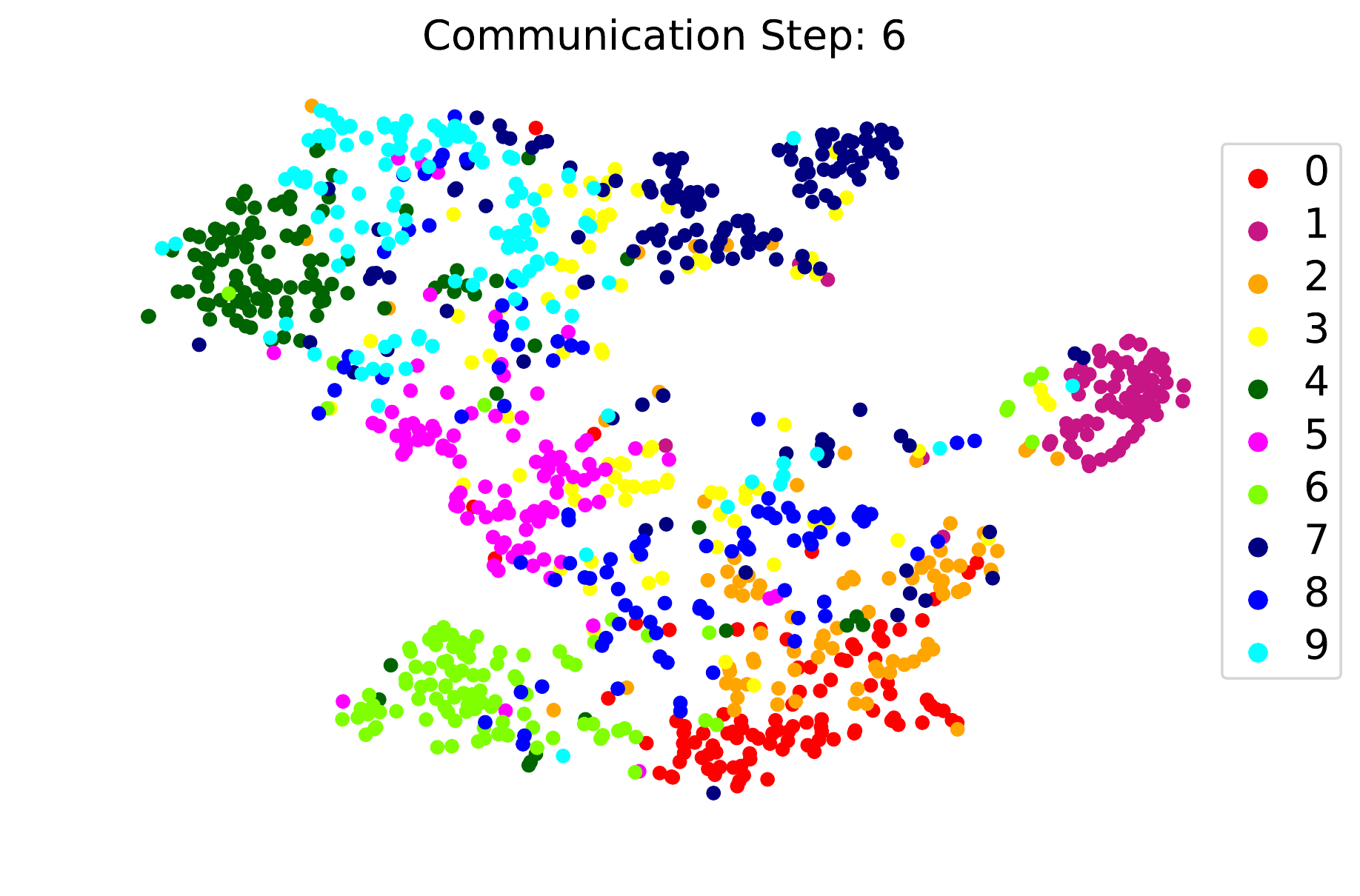}}
			\caption{ Visualization of the learned communications strategies and how they share their beliefs with each other using t-SNE plots.}
			\label{fig:tsne}
		\end{figure}
		
		\noindent{\textbf{Video:}} We have created a     video describing our framework as well as the results of our experiments:
		
		\url{https://youtu.be/j67sy8RK_A4}

		\section{ {Concluding} Remarks}

		{We introduce and analyze a multi-agent image classification framework using {a generalized} policy gradient as the core reinforcement learning technique. The underlying ideas that are discussed in this paper are applicable to the other value-based, policy-based, or  while using novel variance reduction techniques \cite{johnson2013accelerating}. {The problem studied in this paper has a discrete action space within a typically short time horizon. Depending on the problem structure (e.g., in aerial robotic applications), one may prefer a continuous action space. Then, it is straightforward to generalize our methodology to deal with these policies within this framework; for instance, using Gaussian policies \cite{zhang2016learning}.}  
			
			Our extensive simulations suggest that the current models are temporally robust: if we train a model for   time horizon $T=T_1$ and execute the model for $T=T_2>T_1$, the prediction quality using the second model will remain {high}. Also, changing the number of agents will not result in dramatic performance degradation. For example, a model trained with $3$ agents will still produce acceptable outcomes for   problems with $2$ or $4$ agents.}  Due to space limitations, we have not included the related numerical experiments. 
		
	Our numerical experiments have been limited to $2$-D image classification. { However,  this framework can be applied, with minor adjustments, to more realistic scenarios. For instance, as explained in Example \ref{ex:two}, the current framework allows the classification of $3$-D objects using a sequence of intelligently chosen $2$-D observations conducted by moving agents. The optimal (stochastic) movement of the agents around the object could be potentially related to the \emph{next best view} problem \cite{connolly1985determination}. {Moreover, we expect that the sample efficiency of} our methodology can be potentially { enhanced}  {by incorporating the developed optimal movement theories (e.g.{,} see  \cite{esteves2018learning}).} It is {also} an interesting line of research to study the cases where the graph structure dynamically (e.g. randomly) evolves over time.

			\bibliographystyle{IEEEtran}
			
			\bibliography{mybib}

		\end{document}